\documentclass[10pt]{article} 
\usepackage[preprint]{rlc}
\usepackage{preamble}

\usepackage{amssymb}            
\usepackage{mathtools}          
\usepackage{mathrsfs}           
\usepackage{graphicx}           
\usepackage{subcaption}         
\usepackage[space]{grffile}     
\usepackage{url}                
\usepackage{color}
\usepackage{comment}
\usepackage{booktabs}
\usepackage[normalem]{ulem}
\usepackage{multirow}
\usepackage{soul}

\newcommand{\wt}{\bm{\mathrm{w}}}
\newcommand{\zt}{\bm{\mathrm{z}}}

\renewcommand{\paragraph}[1]{{\vspace*{6pt}\noindent\bf{#1 }}}

\definecolor{lightgreen}{RGB}{175,255,175}
\definecolor{lightred}{RGB}{255,150,150}
\definecolor{darkgrey}{RGB}{170,170,170}
\definecolor{lightorange}{RGB}{255,207,158}
\definecolor{cyan}{RGB}{0,255,255}
\definecolor{orange}{RGB}{255,127,80}
\definecolor{darkgreen}{RGB}{50,127,0}
\definecolor{blue}{RGB}{0,0,255}

\newcommand{\SUS}{SUSFAS}

\newcommand{\CUS}{CUSFAS}

\newcommand{\ignore}[1]{\if{0} #1 \fi}
\newcommand{\nignore}[1]{\if{true} #1 \fi}

\theoremstyle{definition}

\theoremstyle{remark}

\renewcommand*{\thefootnote}{\fnsymbol{footnote}}

\title{Stacked Universal Successor Feature Approximators for Safety in Reinforcement Learning}

\author{%
    Ian Cannon\footnotemark[1] \\
    ian.cannon@udri.udayton.edu \\
    University of Dayton Research Institute\\
    \And
    Washington Garcia\footnotemark[1]  \\
    washington.garcia@udri.udayton.edu \\
    University of Dayton Research Institute
    \AND
    Thomas Gresavage\footnotemark[1] \\
    thomas.gresavage@udri.udayton.edu \\
    University of Dayton Research Institute\\
    \And
    Joseph Saurine\footnotemark[1] \\
    joseph.saurine@udri.udayton.edu \\
    University of Dayton Research Institute
    \AND
    Ian Leong\\
    ian.leong@udri.udayton.edu \\
    University of Dayton Research Institute\hspace{1.5em}~
    \And
    Jared Culbertson \\
    jared.culbertson@afrl.af.mil \\
    Air Force Research Laboratory\\
}

\begin{document}

\maketitle

\def\thefootnote{*}
\footnotetext{Indicates equally contributing authors.}
 
\begin{abstract}
Real-world problems often involve complex objective structures that resist distillation into reinforcement learning environments with a single objective. Operation costs must be balanced with multi-dimensional task performance and end-states' effects on future availability, all while ensuring safety for other agents in the environment and the reinforcement learning agent itself. System redundancy through secondary backup controllers has proven to be an effective method to ensure safety in real-world applications where the risk of violating constraints is extremely high. In this work, we investigate the utility of a stacked, continuous-control variation of \textit{universal successor feature approximation} (USFA) adapted for \textit{soft actor-critic} (SAC) and coupled with a suite of secondary safety controllers, which we call \textit{stacked USFA for safety} (SUSFAS). Our method improves performance on secondary objectives compared to SAC baselines using an intervening secondary controller such as a \textit{runtime assurance} (RTA) controller.
\end{abstract}

\section{Introduction}
\label{sec:introduction}

A recent interest in the study of {\em reinforcement learning} (RL) is the ability of learning agents to solve increasingly complex environments yet simultaneously optimize behavior across multiple salient, non-trivial objectives~\citep{yuan_plan4mc_2023,dac}. 
To model these types of problems, recent work has proposed the regime of sequential decision-making tasks with multiple objectives~\citep{sf}, formulated as a linear combination of scalar rewards that can drive a reward signal for continuous iterative-learning algorithms, e.g., \textit{soft actor-critic} (SAC)~\citep{sac}. 
Unfortunately, previous work has not studied the interaction between the aforementioned multi-task formulation and the contextual safety risk of mission-critical problems. In environments that model such problems, there is a consistent threat of intervention (or promise of guidance) from a collection of on-the-loop secondary controllers, which are often unaware of the iterative-learning agent's action policy or value function~\citep{feng2023safety}. 
As a guiding example, we leverage the satellite inspection task originally investigated by \citet{van2023deep} where an inspecting satellite agent (called a ``deputy'') is tasked with inspecting a passive target agent (called the ``chief''). In this problem, human operators expect high performance of an agent in completing an inspection (i.e., inspect more points on the chief), but also place a high value on saving fuel for future missions, alongside the requirement of safe operation through a \textit{runtime assurance} (RTA) secondary controller~\citep{dunlap_rta_2023}. Moreover, the relative importance of these factors might be dynamic, both within a given mission and between missions. 

We study \textit{successor features} (SFs) as they enable direct control over the behavior of an agent by defining mission tasks, represented as composite reward weights~\citep{sf}. 
We expect SF methods to enable better agent generalization over a range of tasks than with SAC alone. We find that stacked SFs lead to agents which effectively encode the behaviors of a secondary safety controller, and likewise serve as efficient primary controllers in safety-critical applications. 

In this work, we build a stacked variation of \textit{universal value function approximation} (UVFA)~\citep{uvfa} for continuous control to predict individualized SFs in safety-critical RL environments. 
We, therefore, call our approach \textbf{Stacked Universal Successor Feature Approximation for Safety (SUSFAS)}. 
While controllability and generalization are important features of SF applications (as introduced by \citet{dayan:sr} and \citet{usf}), this work focuses on the surprising interactions our method has between secondary objectives and intervening controllers. Our contributions are as follows:

\begin{enumerate}
    \item We present a new method of predicting SFs that we call expert stacking, which ensures that each SF is learned independently.
    \item We demonstrate that our RL agents can learn the behaviors of a safety controller and serve as an efficient primary controller in safety-critical applications. Through our ablation studies, we describe the performance gap on secondary objectives in a number of different configurations, including agents acting as so-called generalists and specialists.
    \item We investigate the effects of limiting reward weight ranges during training to evaluate if more specialized agents show improvement on important tasks.

\end{enumerate}
  
We observe that fuel usage (as proxied by $\Delta V$) in mission-critical environments can decrease by up to 18x when trained with an RTA controller when using our method over a SAC primary controller. To encourage reproducibility and advance safety-critical RL, we will publicly release our code.

\section{Background} \label{sec:background}

\label{sec:background:mdp}
We base our formulation on a standard \textit{Markov decision process} (MDP) $M(\St, \A, p, R, \gamma)$ with state space $\St$, action space $\A$, transition probabilities $p(s' \mid s, a)$, a reward function $R\colon \St \times \A \times \St \to \R$, and discount factor $\gamma \in [0,1]$~\citep{puterman}. Based on this standard definition, we first describe \textit{successor features}  (SFs), \textit{universal value function approximation} (UVFA), and their combination as it relates to our technique described later in~\Cref{sec:method}. Afterwards, prior art and relevant techniques involving RL with secondary controllers are discussed. 

\subsection{Demystifying SFs, UVFA, and their combination}

In order to describe USFA, we begin with the following observation: if the reward function decomposes into a linear combination such that $R = \bm\phi^{\top}\wt$, where $\bm{\phi}(s_{t}, a_{t},s_{t+1})\in\R^d$, then by varying $\wt$, we obtain a family of reward functions $R_{\wt}$. Under this observation, the composite rewards $\bm{\phi}$ are known as successor features (SFs)~\citep{dayan:sr}. Subsequently, this definition yields a family of MDPs similarly parameterized by $\wt$. That is, for a given SF $\bm{\phi}\colon \St \times \A \times \St \to \R^d$, a weighting $\wt\in\R^d$ induces an MDP $\M_{\wt}^{\bm{\phi}}(\St,\A,p,\gamma)$ following~\citet{sf} as

\begin{equation} \label{eq:def:sf:mdp}
    \M_{\wt}^{\bm{\phi}}(\St,\A,p,\gamma) \equiv \left \{ M(\St,\A,p,R_{\wt},\gamma) \mid R_{\wt} = \bm{\phi}^{\top}\wt \right \}, 
\end{equation} 

where $R_{\wt}$ and $\bm{\phi}$ are shorthand for $R_{\wt}(s_{t},a_{t},s_{t+1})$ and $\bm{\phi}(s_{t},a_{t},s_{t+1})$, respectively. 

\label{sec:background:sf}
Again following~\citep{sf}, SFs of $(s,a)$ under policy $\pi$ are then defined as
\begin{equation} \label{eq:def:sf:sf}
    \bm{\psi}^{\pi}\equiv\E^{\pi}\left [ \sum_{i=t}^{\infty}\gamma_{t}^{i-t}\bm{\phi}_{t+1} \mid s_{t}=s, a_{t}=a \right ],
\end{equation}
where we omit $\pi$ below when it is implied by the context. 
In short, SFs provide a way to estimate the value of $\pi$ on any task $\wt$ and thus can be deployed in the \textit{policy evaluation} step of {\em dynamic programming} (DP) methods~\citep{sutton1998}. \citet{sf} later use \textit{generalized policy improvement} (GPI) to find an improved policy from the set of policies with known action-value function approximations provided by SFs. In this way, SFs with GPI can provide coverage of the entire value space for a given policy, whereas coverage of the entire policy space is impossible to achieve with SF and GPI alone.

In contrast, UVFA seeks to ``represent a large set of optimal value functions by a unified function approximator that generalizes over both states and goals''~\citep{uvfa}. Since UVFA allows us to condition the value function on the goals, we are able to combine this with SF and DP methods to generate goal-conditioned policies and compare the performance of any policy on any task. Hence, \textit{Universal successor feature approximation} (USFA) seeks to provide coverage of policy space not provided by SFs alone. Even though our SUSFAS method uses UVFA-inspired blocks in the architecture, we instead use them to predict SFs. Therefore, we refer to these blocks as \textit{successor feature approximators} (SFAs) to avoid confusion when describing our method.

\subsection{Related Work} \label{sec:background:related-work}

To date, secondary controllers are a common technique for providing redundancy and safety to RL agents in mission-critical problems. This ensures the safe operation of complex systems as they perform difficult tasks across a wide variety of domains and, therefore, has become the industry standard in many high-value control domains~\citep{fuller_rta_2020}. RTA solutions have demonstrated value in high-performance aerospace systems such as Auto GCAS~\citep{hobbs2023runtime}, aircraft swarms~\citep{schierman2020runtime}, satellite operations~\citep{mote2021natural}, wireless sensor networks \citep{wu2010run}, and software verification~\citep{kim2004java}. Detailed primers for this area of research are provided by~\citet{hobbs2023runtime} and~\citet{fuller_rta_2020}. In general, an RTA solution implements a secondary controller that is agnostic to the underlying policy or design of the RL agent. Instead, these techniques rely on on-the-loop intervention which activates before an agent takes any action toward an unsafe state. In terms of our guiding example, the RTA will activate according to pre-specified bounding constraints (when an agent attempts to maneuver toward any state that approaches the out-of-bounds limit of the mission), thereby correcting any unsafe action before the state becomes unsafe. 

The work of~\citet{usfa} extends that of~\citet{sf} by combining SF\&GPI with UVFAs \citep{uvfa} to achieve a coverage of policy space. This work uses a form of Watkins Q-learning \citep{Watkins:89} for the SF\&GPI step and does not directly address the continuous control problem. Our strategy is quite similar to the strategy adopted by \cite{mozifian_USFAforCC_2021} with the exception that we do not use approximators to estimate $\bm{\tilde{\mathrm{w}}}$ and $\bm{\tilde{\phi}}$. Theoretically, the results ought to be applicable to any RL algorithm, including actor-critic methods, and thus, we are motivated to empirically test the theory in the continuous domain. It is unclear what advantage the methods of SF\&GPI + UVFA afford the algorithm. The SF framework learns on the feature vector $\bm{\phi}$, but it is not well-understood if generalization is still possible when learning on $r_{\wt}$ directly. Taking inspiration from SF, UVFA, and USFA, we rework USFA for safety applications. As the layers are not stacked in this formulation, our work refers to the USFA-inspired architecture as \textit{collapsed USFA for safety} (CUSFAS) compared to our \textit{stacked USFA for safety} (SUSFAS), described in detail in~\Cref{sec:exp:detail:stacking}.

\section{Stacked Universal Successor Feature Approximation for Safety}
\label{sec:method}

\begin{figure}[t]
\begin{subfigure}{1.0\textwidth}
    \centering
    \tikzset{
    block/.style = {draw, fill=white, rectangle, minimum height=3em, minimum width=3em},
    tmp/.style = {coordinate},
    sum/.style = {draw,circle,append after command={
        [shorten >=\pgflinewidth, shorten <=\pgflinewidth,]
        (\tikzlastnode.north) edge (\tikzlastnode.south)
        (\tikzlastnode.east) edge (\tikzlastnode.west)
        }},
    line/.style = {draw, -latex', shorten <=1bp, shorten >=1bp}
}

\definecolor{stop}{HTML}{0ACA50}
\definecolor{sbot}{HTML}{0AA070}
\definecolor{atop}{HTML}{A97FBA}
\definecolor{abot}{HTML}{A9379A}
\definecolor{wtop}{HTML}{FACA50}
\definecolor{wbot}{HTML}{FAA070}
\definecolor{otop}{HTML}{4AF0FF}
\definecolor{obot}{HTML}{4A8FFF}

\tikzset{
    uvfa-block/.pic={

    \node[block, fill= gray!50!white, minimum width= 21em, minimum height = 15em] (uvf1) at (4.9,-1.75) {};
    \node[block, fill = stop, opacity=.66] (w1) at (0,0) {};
    \node[block, fill = atop, opacity=.66, below of = w1, node distance = 5em] (a1) {};
    \node[block, fill = wtop, opacity=.66, below of = a1, node distance = 5em] (s1) {}; 
    \node[block, shading = axis, top color = stop, bottom color = sbot, right of = w1, node distance = 7em, minimum width = 5em] (wencode) {};
    \node[block, shading = axis, top color = atop, bottom color = abot, right of = a1, node distance = 7em, minimum width = 5em] (aencode) {};
    \node[block, shading = axis, top color = wtop, bottom color = wbot, right of = s1, node distance = 7em, minimum width = 5em] (sencode) {};
    \node[sum, right of = aencode, node distance = 10em] (c1) {};
    \node[block, shading = axis, top color = otop, bottom color = obot,right of = c1, node distance = 5em] (o1) {};

    \draw[->, thick] (w1) to (wencode);
    \draw[->, thick] (a1) to (aencode);
    \draw[->, thick] (s1) to (sencode);
    \draw[-,thick, shorten >=2bp] (wencode.east) -| (c1.north);
    \draw[-,thick, shorten >=2bp] (aencode.east) to (c1.west);
    \draw[-,thick, shorten >=2bp] (sencode.east) -| (c1.south);
    \draw[->, thick, shorten >=2bp, shorten <=2bp] (c1) to (o1);
    }
}

\newcommand{\centerbase}{-4ex}

\begin{tikzpicture}[transform shape, scale=.75]
    \pic[transform shape, scale=0.3, anchor=center] (block1) at (0,0) {uvfa-block};
    \pic[transform shape, scale=0.3, anchor=center] (block2) at (0,-2) {uvfa-block};
    \pic[transform shape, scale=0.3, anchor=center] (block3) at (0,-4) {uvfa-block};
    \pic[transform shape, scale=0.3, anchor=center] (blockd) at (0,-7) {uvfa-block};

    \node[block, fill= gray!50!white, thick, below left = 1em and 12em of block1o1, minimum height=19em, minimum width=30em] (uvfa) {};
    \node[block, very thick, dashed, fill opacity=0, above left = 1em and 6em of uvfa.north west, anchor=north west, minimum height=21em, minimum width=37em] (uvfa_border) {};

    \draw[-, very thick, dashed] ([yshift=.5em]block1uvf1.north) -- ([yshift=1.5em]block1uvf1.north) -| (uvfa_border.north); 

    \coordinate (centerdots) at ($(block3uvf1.south)!.50!(blockduvf1.north)$);
    \coordinate (sumpos) at ($(block2uvf1.south)!.50!(block3uvf1.north)$);

    \node at ([yshift=4pt]centerdots) {$\vdots$};
    \node[sum, right = 5em of sumpos] (c2) {};
    \node[block, fill = otop, opacity=.66, right = 2em of c2] (o2) {$\bm{\tilde\psi}$};

    \draw[-, thick, shorten >=2bp] (block1o1) -| (c2.north);
    \draw[-, thick, shorten >=2bp] (block2o1) -| (c2.north);
    \draw[-, thick, shorten >=2bp] (block3o1) -| (c2.south);
    \draw[-, thick, shorten >=2bp] (blockdo1) -| (c2.south);
    \draw[->, thick, shorten >=2bp, shorten <=2bp] (c2) -- (o2);

    \draw[-, very thick, dashed] ([xshift=-2em, yshift=.5em]block1uvf1.north west) -- ([xshift=.5em, yshift=.5em]block1uvf1.north east) -- ([xshift=.5em, yshift=-.5em]block1uvf1.south east) -- ([xshift=-2em, yshift=-.5em]block1uvf1.south west) -- ([xshift=-2em, yshift=.5em]block1uvf1.north west);

    \node[block, shading=axis, top color = stop, bottom color = sbot, below right = .5em and 2em of uvfa.north west, align = center] (se)
    {\bf State Encoder \\[1em]
        \tikz[baseline=\centerbase]{\node[block, fill opacity=0, text opacity=1]{$\tilde \varsigma_{0,0}$}}
        $\to \cdots \to$ 
        \tikz[baseline=\centerbase]{\node[block, fill opacity=0, text opacity=1, baseline=\centerbase]{$\tilde \varsigma_{0, h_\varsigma}$}}};

    \node[block, shading=axis, top color = atop, bottom color = abot, below = 5bp of se, align = center] (ae)
    {\bf Action Encoder \\[1em]
        \tikz[baseline=\centerbase]{\node[block, fill opacity=0, text opacity=1]{$\tilde \alpha_{0,0}$}}
        $\to \cdots \to$ 
        \tikz[baseline=\centerbase]{\node[block, fill opacity=0, text opacity=1, baseline=\centerbase]{$\tilde \alpha_{0, h_\alpha}$}}};
    
    \node[block, shading=axis, top color = wtop, bottom color = wbot, below = 5bp of ae, align = center] (we)
    {\bf Task Weight Encoder \\[1em] 
        \tikz[baseline=\centerbase]{\node[block, fill opacity=0, text opacity=1] {$\tilde \omega_{0,0}$}}
        $\to \cdots \to$ 
        \tikz[baseline=\centerbase]{\node[block, fill opacity=0, text opacity=1, baseline=\centerbase]{$\tilde \omega_{0, h_\omega}$}}};    

    \node[block, fill = wtop, opacity=.66, below left = -33pt and 3.5em of we, align=center] (w) {$\wt$};
    \node[block, fill = atop, opacity=.66, below left = -33pt and 3.5em of ae] (a) {$a_t$};
    \node[block, fill = stop, opacity=.66, below left = -33pt and 3.5em of se] (s) {$s_t$};

    \node[sum, right = 2em of ae] (uvfa-sum) {};
    \node[block, shading=axis, top color = otop, bottom color = obot, right = 2em of uvfa-sum, align=center] (out)
    {\bf Output Encoder \\[1em]
        \tikz[baseline=-1ex]{\node[block, fill opacity=0, text opacity=1]{$\tilde\psi_{0,0}$}}
        $\to \cdots \to$
        \tikz[baseline=-1ex]{\node[block, fill opacity=0, text opacity=1]{$\tilde\psi_{0,h_{\psi}}$}}};

    \draw[->, thick, shorten >=2bp] (w) to ([yshift=17.5pt]we.south west);
    \draw[->, thick, shorten >=2bp] (a) to ([yshift=17.5pt]ae.south west);
    \draw[->, thick, shorten >=2bp] (s) to ([yshift=17.5pt]se.south west);

    \draw[-, thick, shorten >=2bp] (we) -| (uvfa-sum);
    \draw[-, thick, shorten >=2bp] (ae) -- (uvfa-sum);
    \draw[-, thick, shorten >=2bp] (se) -| (uvfa-sum);
    \draw[->, thick, shorten >=2bp] (uvfa-sum) -- (out);
    \draw[->, thick] (out.east) -- ([xshift=3em]out.east);


    \node (alabel) at ([yshift=-3em]uvfa.south|-current bounding box.south) {(a) SUSFAS Architecture Block};
    \node at (blockdo1|-alabel) {(b) SUSFAS Architecture Stack};

\end{tikzpicture}
    \phantomcaption
    \label{fig:vfd_sac_block}
\end{subfigure}
\begin{subfigure}{0.0\textwidth}
    \phantomcaption
    \label{fig:usfa_stack}
\end{subfigure}
\caption{Our proposed architecture, with (a) a detailed SUSFAS architecture block and (b) the overall SUSFAS architecture stack. The variables $\bm{\tilde{\varsigma}}_{0,h_{\varsigma}}$, $\bm{\tilde{\alpha}}_{0,h_{\alpha}}$, $\bm{\tilde{\omega}}_{0,h_{\omega}}$, and $\bm{\tilde{\psi}}_{0,h_{\psi}}$ indicate the feature embedding of the states, actions, task weights, and combined representation after $h_{\varsigma}$, $h_{\alpha}$, $h_{\omega}$, and $h_{\psi}$ hidden layers respectively. The output corresponds to the \textit{successor feature approximation} (SFA) for the $0$-th successor feature, $\tilde{\psi}_0$. The $\oplus$ symbol indicates concatenation. Our SUSFAS Architecture Stack (b) shows how the $d$-many SFA blocks are stacked and concatenated to form the successor feature vector $\bm{\tilde{\psi}}$.}
\label{fig:vfd_sac_block}
\end{figure}
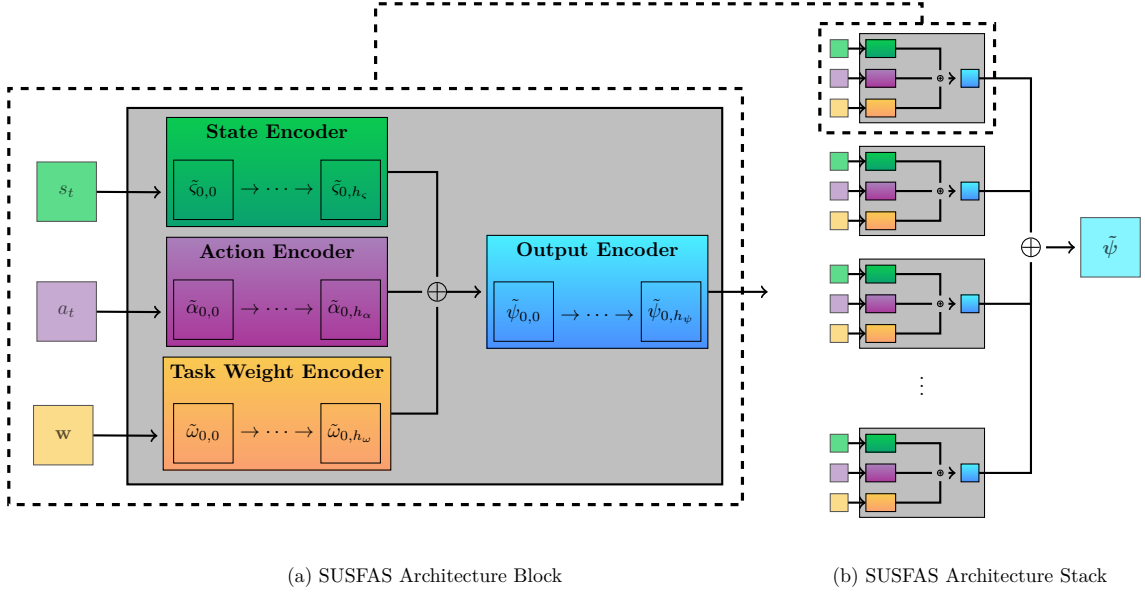

In this section, we describe our \textit{stacked universal successor features for safety} (SUSFAS) method. To reformulate USFA to continuous control, the Q-network is replaced by a network trained to predict the successor representation instead of the scalar Q-value. The task-weight sampling step used by~\citep{usfa} is left unchanged. The sampling of alternative task weightings is itself optional, but it is to be expected that generalization suffers by virtue of only learning about the target task $\wt$. Therefore we chose to keep this feature of the original USFA paper. The burden of determining the next action $a^{\prime}$ according to $\argmax_{b}\max_{i}\bm{\tilde{\psi}}(s,b,\zt_{i})$ is similar in the continuous case, with the only difference that $b_{i}\sim\pi(s,\zt_{i})$ is an action sampled from the task-conditioned policy function approximator instead of using a lookup table.

There are some notable differences where we diverge from the inspirational work described in \Cref{sec:background}. Our algorithm uses an array of $d$-many UVFA-inspired \textit{successor feature approximators} (SFAs), each trained to predict a single successor feature, decomposed according to to~\Cref{eq:def:sf:mdp}. Each SFA has a separate encoder for observations, actions, and task weights. Hence we call this the \textit{stacked USFA} (SUSFA) architecture. We use GPI and the policy sampling methodology of~\citep{usfa} to train the critic and select actions. Furthermore, our actor does not use separate encoders but makes predictions on a vector of observations concatenated with the task weights. This is in contrast to a \textit{collapsed USFA} (CUSFA) architecture, under which we consider \citet{usfa} to fall as it uses the same observation and weight encoders for every SF.

\Cref{fig:vfd_sac_block}a shows how a single SFA block works. For the approximator assigned to the $i$-th successor feature, the state observations are passed through an encoder with $h_{\varsigma}$ hidden layers. The intermediate representation after the $j$-th hidden layer is denoted $\bm{\tilde{\varsigma}}_{i,j}$. Here, and throughout, we use a tilde to denote a value generated by some function approximator rather than an exact known value. Similarly, for the action encoder with $h_{\alpha}$ hidden layers, the representation in the $i$-th block after $j$ hidden layers is $\bm{\tilde{\alpha}}_{i,j}$. Finally, the task weights are encoded by $h_{\omega}$ hidden layers and denoted as $\bm{\tilde{\omega}}_{i,j}$ after $j$ hidden layers in the $i$-th SFA block. These representations are concatenated together before being fed as the input to the output encoder with $h_{\psi}$ hidden layers, which produces the $j$-th intermediate representations denoted by $\bm{\tilde{\psi}}_{i,j}$. The final output of this approximator block is trained on the {\em mean-squared Bellman error} (MSBE) of the corresponding component of the successor representation $\psi_i$.

In \Cref{fig:usfa_stack}, we show how the scalar outputs of $d$ of these individual SFA blocks are concatenated to form the full predicted successor feature vector $\bm{\tilde{\psi}}$, we call this operation \textit{expert stacking} (ES). Each of these $\psi$-networks plays the role of the $Q$-networks in SAC (we refer the reader to \citep{sac} for details on SAC). Thus, we have $d$-many $\psi$-networks $\{\psi_m\}_{m \in \{1,2\}}$ with assoicated parameters $\xi_m$ training independently on the same data. Each $\psi_m$ has a corresponding target $\psi$-network $\bar\psi_m$, which is initialized with $\bar{\xi}_m=\xi_m$ to have the same initial parameters. After $k$ updates to the networks for $\psi$-networks, we perform a Polyak update to the parameters of their corresponding target networks according to a hyperparameter $\eta$, which describes the magnitude of the update (see \ref{fig:app:hparams}). In other words, for the target networks after $k$ steps we set $\bar{\xi}_{m,t+1} = \eta\cdot\bar{\xi}_{m,t} + (1 - \eta)\cdot\xi_{m,t}$. The minimum $Q$-values from the target networks, defined as $\bar{Q}^\prime=\min_{m}\bar{Q}_{m}$, is used to update the actor network using the loss for SAC in \Cref{eq:vfd_sac_loss}. It is important to note that the scalar $Q$-value prediction, rather than $\bm{\tilde{\psi}}$, is used to update the actor network. The overall SAC loss is defined as

\begin{align}\label{eq:vfd_sac_loss}
    \mathcal{L}_{\pi} &= \sum_{i=0}^{n_{\zt}}\tau\log\left[\pi_{i}\mkern-3mu\left(a_{i} | s_{t}\right)\right]- \bar{Q}^\prime\left(s_t,a_i\right),\nonumber \\
    \mathcal{L}_{\psi_{m}} &= \sum_{i=0}^{n_{\zt}}\left\|\bm{\tilde{\psi}}_{m}\left(s_t,a_{i}\right)-\left(\bm{\phi}_t + \gamma\bm{\tilde{\psi}}_{m}\left(s_{t+1},a_{i}^\prime\right)\right)\right\|^2, \\
    \mathcal{L}_{\alpha} &= \sum_{i=0}^{n_{\zt}}-\tau\log\left[\pi_{i}\mkern-3mu\left(a_{i} | s_t\right)\right]-\tau\bar{\mathcal{H}}, \nonumber
\end{align}

where we take the weights as $\wt$ whenever $i=0$ and $\zt_i\thicksim\mathcal{D}$ otherwise (note that $\mathcal{D}$ may be an arbitrary distribution used to generate $\zt$), we denote by $a_{i}^\prime$ the action for task alternative $i$ at time $t+1$, we can obtain $\bm{\phi}_{t}$ from the observed reward such that $r_{t}=\bm{\phi}_{t}^\top{\wt}$, the SAC \textit{temperature} parameter $\tau$ controls the degree of entropy regularization, and $\mathcal{\bar{H}}$ is the target entropy which is usually taken to be $\dim\mathcal{A}$ for SAC~\citep{sac}. 

We adopt the GPI training regimen of~\citep{usfa} wherein each timestep we generate $n_{\zt}$ task weights as alternatives to $\wt$ and choose the action having the highest $Q$-value on the true task $\wt$ (note the use of $Q$). The minimal $Q$-value from the twin $\psi$-networks is used for this assessment. These alternative task weights are stored in the replay buffer and used to generate new actions during loss calculation so that our actor and critic networks learn $n_{\zt}+1$ tasks simultaneously. In our case, we construct a multivariate Gaussian around $\wt$ to generate each $\zt$.

\section{Experiments \& Results}
\label{sec:exp}

\subsection{Environments} 
\label{sec:exp:setup:env}

In our experiments, we focus on two environments selected for their requirement to solve a challenging secondary objective while also solving a relatively simple primary objective---easy to learn, difficult to master. We primarily leverage the environment proposed by~\citet{van2023deep}, \textbf{Inspection3D}, which models a translational $3$-dimensional satellite inspection problem. In this case, the agent is subject to a challenging secondary objective of reducing fuel usage while solving a primary objective of inspecting some stationary (from a relative reference frame) satellite and avoiding a crash. The environment includes a pre-built RTA secondary controller~\citep{dunlap_rta_2023} which allows us to perform several ablation studies similar to~\citet{hamilton_ablation_2022}. A key characteristic of this environment is that it is relatively straightforward to achieve near-perfect inspection success using many common algorithms, but it is much more challenging to do so while maintaining low fuel usage. See the original implementation by~\citet{van2023deep} for details on the dynamics and illumination models along with the description of RTA implementation by~\citet{dunlap_rta_2023}.

In addition, we evaluate our algorithm on a simplified environment, Multi-Objective Lunar Lander (denoted Lunar Lander in figures), which has fuel usage and crash avoidance tasks~\citep{felten_toolkit_2023}. In order to test agents in the context of secondary controller intervention, we modify the environment to include an RTA-inspired secondary controller that assists the agent in safely landing, described in~\Cref{alg:lunar_controller}. The controller supports the agent to be horizontally close to the goal ($\left | \varphi_{x} \right | \leq 0.3$), prevent steep body angles ($\left | \delta \right | > 0.4$), and control vertical velocity ($\left ( -0.8 * \dot\varphi_{y} > \varphi_{y} \right ) \land \left ( \varphi_{x} > 0.03 \right )$). The lunar lander environment is considered particularly easy to solve by contemporary RL algorithms but difficult to refine due to the necessity for minimal fuel usage. 

In our evaluation, all agents are trained to convergence and are capable of solving the environments, scoring similarly across the board for the primary objectives (e.g., inspected points reward in Inspection3D, crash or land reward for Lunar Lander). Hence, we primarily focus on plotting the difficult secondary objective of fuel usage.

\defcitealias{sac}{HZH18}
\defcitealias{dunlap_rta_2023}{DWH23}

\begin{figure}[t]
  \begin{minipage}[c]{.49\linewidth}
    \centering
    \begin{algorithm}[H]
    \caption{Lunar Lander Secondary Controller}
    \label{alg:lunar_controller}
    \begin{algorithmic}[1]
    \Require Main engine throttle $a_{main}$, side engine throttle $a_{side}$, and craft's position ($\varphi_{x}$, $\varphi_{y}$), velocity ($\dot\varphi_{x}$, $\dot\varphi_{y}$), body angle $\delta$, and angular velocity $\dot\delta$.
    \For{step $i$ in episode}
        \If{$\left | \varphi_{x} \right | > 0.3$ \textbf{or} $\left ( -0.8 * \dot\varphi_{y} > \varphi_{y} \right ) \land \left ( \varphi_{x} > 0.03 \right )$ \textbf{or} $\left | \delta \right | > 0.4$}
            \State $a_{side} \gets (10 * \delta) + (3 * \dot\delta) - (2 * \varphi_{x}) - (4 * \dot\varphi_{x})$
            \If{$\dot\varphi_{y} > -0.001$}
                \State $a_{main} \gets 0$
            \Else
                \State $a_{main} \gets 1$
            \EndIf
        \EndIf
    \EndFor
    \end{algorithmic}
    \end{algorithm}
    \end{minipage}\hfill
  \begin{minipage}[c]{.49\linewidth}
    \centering
    \footnotesize	
    \begin{tabular}{lcc}
    \toprule
    Policy & \begin{tabular}{@{}c@{}}Implementation \\ Reference\end{tabular} & \# Parameters \\
    \midrule
    SAC Spec. & \citetalias{sac} & 355,082 \\
    SAC Gen. & \citetalias{sac} & 355,082 \\
    SAC-S Spec. & \citetalias{dunlap_rta_2023} & 356,362 \\
    SAC-S Gen. & \citetalias{dunlap_rta_2023}  & 356,362 \\
    SUSFA & \Cref{sec:method} & 7,522,070 \\
    SUSFAS w/o $R$ & \Cref{sec:method} & 7,522,070 \\
    SUSFAS & \Cref{sec:method} & 9,390,106 \\
    CUSFAS & \Cref{sec:exp:detail:stacking} & 1,939,482 \\
    \bottomrule
    \end{tabular}
    \captionof{table}{Parameter counts for studied policy architectures.}
    \label{tab:params}
  \end{minipage}
\end{figure}

\subsection{Hyperparameters and Architectures}
\label{sec:exp:setup:params}

In order to investigate the structural advantages of our modified successor and network structure, we compare agents structured and trained across different combinations of architectures, agent types, SF weight ranges, and safety contexts. These experimental variables are summarized as follows:

\paragraph{Agent types.} Based on existing literature, we distinguish between two primary agent types. The first, so-called \textit{specialists}, have been trained to specialize on a fixed set of reward weights $\bm{\mathrm{w}}$. In practice, we lock the reward weights for specialists to $1.0$. Reward weights for specialists are not normalized to sum to $1$. As a complement to specialists, we train \textit{generalists}, which may have their reward weights tuned within a range of values during evaluation time. In the generalist case, reward weights are normalized to sum to $1$ before feeding them to the task encoder.

\paragraph{Architectures.} We train agents using the SAC algorithm and treat this as a baseline implementation since it is the closest comparable architecture to our proposed SUSFA method. Our proposed SUSFA agents are trained using the architecture described in~\Cref{sec:method}. To generate SAC generalists, we input the modified weights as part of the SAC agent's observation, which ensures a fair comparison to SUSFA generalists. Conversely, we train a variation of generalist and specialist SUSFA agents that have, during training, either observed a weight range (e.g., $\wt \in [0,1]^d$, inducing a SUSFA generalist) or a single value for all weights (e.g., $\wt = \mathbf{1}$, inducing a SUSFA specialist). In addition to these SUSFA variants, we train CUSFA, which examines the effect of collapsing the stack of expert networks. 

\paragraph{Secondary intervention.} To examine the interaction of safety-oriented secondary intervention with generalist architectures (e.g., SAC and SUSFAS generalists, as described earlier), we ablate between having RTA on and off during training. In the case that RTA is on (off), we denote agents as SAC-S (SAC) or SUSFAS (SUSFA), respectively.

Across all experiments, we conduct training and evaluation over six seeds, with evaluation taking place across ten episodes for each fixed seed, totaling $60$ combinations of seed and episodes for each technique. Code for SUSFAS and subsequent experiments will be released publicly to encourage reproducibility. 

\subsection{Experimental Details}

We present our findings relating to SUSFAS by first stating a motivating research question (RQ), following up with experimental details, and then concluding with the research answer (RA). 

\subsubsection{Investigation of expert stacking}
\label{sec:exp:detail:stacking}

\begin{table}[t]
\centering
\footnotesize
\begin{tabular}{@{}lcccccccc@{}}
                                      & \multicolumn{4}{c}{$\Delta V$ Usage}                                         & \multicolumn{4}{c}{Inspection Reward}               \\ \midrule
\multicolumn{1}{l|}{Experiment Name}  & 2.5M       & 5M         & 7.5M             & \multicolumn{1}{c|}{10M}        & 2.5M      & 5M              & 7.5M      & 10M       \\ \midrule
\multicolumn{1}{l|}{SUSFAS}             & $21.41$    & $21.79$    & $\mathbf{12.30}$ & \multicolumn{1}{c|}{$29.47$}    & $1.99$    & $\mathbf{1.99}$ & $1.97$    & $1.77$    \\
\multicolumn{1}{l|}{}                 & $\pm17.39$ & $\pm35.56$ & $\pm6.15$        & \multicolumn{1}{c|}{$\pm41.89$} & $\pm0.00$ & $\pm0.00$       & $\pm0.13$ & $\pm0.50$ \\
\multicolumn{1}{l|}{CUSFAS} & $48.84$    & $37.40$    & $40.79$          & \multicolumn{1}{c|}{$48.35$}    & $1.57$    & $1.55$          & $1.55$    & $1.56$    \\
\multicolumn{1}{l|}{}                 & $\pm33.66$ & $\pm19.19$ & $\pm14.76$       & \multicolumn{1}{c|}{$\pm22.56$} & $\pm0.60$ & $\pm0.62$       & $\pm0.62$ & $\pm0.61$ \\ \bottomrule
\end{tabular}
\caption{RQ1 -- The effect on fuel usage and inspection reward in Inspection3D during evaluation episodes when stacking SF networks (SUSFAS) or collapsing them (CUSFAS). Our proposed stacking approach leads to an improved fuel usage envelope.}
\label{tab:stacking}
\end{table}

\paragraph{RQ1:} \textit{Expert stacking} (ES) forces agents to learn each \textit{successor feature} SF independently, where gradients from one SF do not influence other SF networks. How does expert stacking compare to a more traditional structure with a collapsed network to predict SFs? 

The work of previous groups~\citet{dac} and~\citet{sfgpixfer} to bring SF representations into the continuous domain has employed the construction of an expert repertoire known as ``expert stacking''~\citep{sf,uvfa}. In this case, a layer or group of layers is specialized to a single task (on the input side of UVFA) or single policy (on the output side of SFs \& GPI). The USFA framework avoids the need to build an expert network for each task by combining the experts into flat layers of the final network. However one could leave these ``expert layers'' separated, i.e. stacked. 

We explore the difference between the pure ``collapsed'' USFA architecture (with a necessary action encoder for dimensional compliance) to the fully ``stacked'' architecture. The expectation is that the greater number of parameters in the stacked architecture would enable a more rich representation as well as reduce the burden of any one level in the stack to need to generalize. To investigate this trade-off, we compare the mean and standard deviation of Inspection3D fuel usage on evaluation episodes at certain timestep thresholds, recorded in~\Cref{tab:stacking}, and measure the number of parameters for every investigated architecture in~\Cref{tab:params}. As evidenced by~\Cref{tab:stacking}, the stacked variant (\SUS) achieves a consistently lower fuel usage (measured by $\Delta V$ usage) compared to the collapsed variant (\CUS) on evaluation episodes. This is despite a similar inspection reward (i.e., the primary objective) being within standard deviation for both methods, showcasing that separating the SF representations for each expert is inherently valuable for encoding signals from a secondary objective. At higher timesteps (over 7.5M), \SUS~diverges slightly for fuel usage at the cost of lower inspected points, meanwhile still outperforming \CUS. This may demonstrate over-fitting due to the increased network size (\Cref{tab:params}). 

\paragraph{RA1:} From~\Cref{tab:stacking}, our SUSFAS method using expert stacking outperforms CUSFAS by using 3x less $\Delta V$ in the Inspection3D environment.

\subsubsection{Ablation of fuel usage in mission-critical environment}
\label{sec:exp:detail:safety}

\paragraph{RQ2:} 
RTA presents an intervening secondary controller that activates when some safety-critical condition is met.
While our method allows agents to learn successor features independently per reward component, is the presence of RTA required to see improvement in our method?

\begin{figure}[t]
     \centering
     \begin{subfigure}[t]{0.49\textwidth}
         \centering
         \includegraphics[scale=0.98]{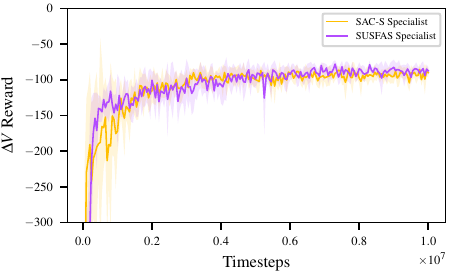}
         \caption{LunarLander Fuel Reward}
         \label{fig:RQ2.1.1}
     \end{subfigure}
     \hfill
     \begin{subfigure}[t]{0.49\textwidth}
         \centering
         \includegraphics[scale=0.98]{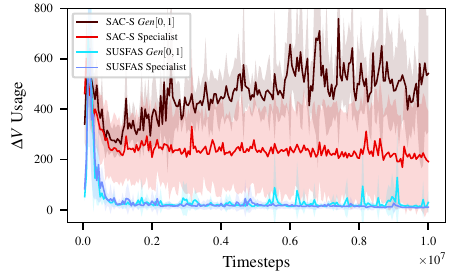}
         \caption{Inspection 3D $\Delta V$ usage}
         \label{fig:RQ2.1.2}
     \end{subfigure}
    \caption{RQ2 -- Investigating effect of RTA on for (a) total fuel usage on Lunar Lander, and (b) $\Delta V$ usage on Inspection3D with specialist and generalist SAC-S and SUSFAS agents. Shaded regions denote the 95\% confidence interval. }
    \label{fig:RQ2.1}
\end{figure}

Given the ability for stacked variants to outperform collapsed variants, we investigate the ability of our \SUS~algorithm to encode the functionality of an RTA controller. In this scenario, there are three cases: RTA off (e.g., SAC or SUSFA), RTA on with a penalty for activating RTA (SAC-S or SUSFAS), and RTA on \textit{without} a reward penalty for activating RTA (SAC-S w/o $R$ (Reward), SUSFAS w/o $R$). With the \textit{RTA off} scenario as our control, we compare the behavior of the algorithms to respond to RTA. By incorporating a penalty for activating RTA into the reward signal, the agent must learn to avoid risky regions of behavior space thus ensuring redundancy of safety systems. First, we present results when RTA is on for all agents, shown in~\Cref{fig:RQ2.1.1} and~\Cref{fig:RQ2.1.2} for Lunar Lander and Inspection3D, respectively. We measure the average fuel usage on evaluation episodes at every $50k$ timesteps of training, and use shaded regions to denote the 95\% confidence interval. Depending on the environment, the presence of an RTA secondary controller can have a major impact on the agent performance. For example, RTA has a minimal impact on the Lunar Lander environment (\Cref{fig:RQ2.1.1}), showcasing similar fuel reward for SUSFAS and SAC-S specialists, meanwhile on Inspection 3D, the same agent architectures in~\Cref{fig:RQ2.1.2} diverge in performance. For example, SAC-S specialists (red) only achieves at best an average fuel usage of $191.82 \pm 200.40$, where SUSFAS specialists (dark blue) reduce this to an average of $10.88 \pm 13.49$ at 10M. For generalists, we see that SAC-S (dark red) diverges after 1M timesteps and fails to recover, whereas SUSFAS (cyan) maintains low fuel usage after convergence (down to $29.47 \pm 41.89$ at 10M).  

\begin{figure}[t]
  \begin{minipage}[c]{.49\linewidth}
     \centering
     \includegraphics[scale=0.98]{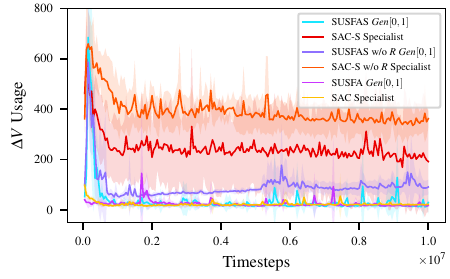} 
    \captionof{figure}{RQ2 -- Inspection3D $\Delta V$ usage with RTA ablations. }
    \label{fig:RQ2.2new}
  \end{minipage}\hfill
  \begin{minipage}[c]{.49\linewidth}
    \centering
    \footnotesize	
    \begin{tabular}{lc}
    \toprule
    Approach & \begin{tabular}{@{}c@{}}Normalized \\ AUC\end{tabular} \\
    \midrule
    SAC-S $Gen[0,1]$ & 1.00 \\
    SAC-S Specialist & 0.51 \\
    SUSFAS $Gen[0,1]$ & 0.04 \\
    SUSFAS Specialist & 0.03 \\
    \midrule
    SAC $Gen[0,1]$ & 0.06 \\
    SAC Specialist & 0.01 \\
    SUSFA $Gen[0,1]$ & 0.01 \\
    SUSFA Specialist & \textbf{0.00} \\
    \bottomrule
    \end{tabular}
    \captionof{table}{RQ2 -- Ablation with/without RTA (SAC-S/SAC, SUSFAS/SUSFA).}
    \label{tab:RQ2.2new}
  \end{minipage}
\end{figure}

Apart from when RTA is on, we also ablate the performance of agents between the three scenarios described in the prequel, shown in~\Cref{fig:RQ2.2new}, primarily for SAC-type specialists and SUSFA-type generalists. It can be observed that, regardless of technique, incorporating an RTA reward penalty encourages better worst-case behavior, visualized by cyan and purple lines (SUSFAS) or red and orange lines (SAC-S). Violet and yellow lines, which demonstrate RTA off (SUSFA and SAC, respectively), show that in the absence of a safety controller, both techniques can perform similarly. It is when RTA is enabled (e.g., cyan, red) that SUSFAS demonstrates better performance, indicating that in a safety-critical context,  the stacked expert networks may be more receptive to a secondary controller due to enhanced encoding ability. To provide a fine-grained comparison of the performance between RTA on and off for all agent types, we calculate the trapezoidal rule area-under-curve (AUC) for the average $\Delta V$ usage across checkpoints. This quantity is shown in~\Cref{tab:RQ2.2new} for both specialist and generalist agents (lower is better), showcasing that SUSFA specialist variants provides the best normalized AUC for $\Delta V$ usage when RTA is off (bold) followed closely by SAC specialists. Even though SUSFAS specialists are an unlikely use case of the SUSFAS architecture, it remains competitive with SAC (SUSFAS AUC$=0.06$, SAC AUC$=0.01$). Likewise, SUSFA-type generalists consistently outperform SAC baselines when RTA is on.

\paragraph{RA2:} We show with~\Cref{fig:RQ2.2new} and~\Cref{tab:RQ2.2new} that, when RTA is off, our SUSFA generalists are comparable to baseline SAC specialists according to the agent's seconary objective (fuel usage). However, when RTA is turned on, SUSFAS generalists significantly outperform both SAC-S generalists and specialists.

\subsubsection{Investigation of safe agent controllability}
\label{sec:exp:detail:control}

\paragraph{RQ3:} Reward weights, $\bm{\mathrm{w}}$, present an interesting new dimension by which to train agents. What effect does altering the $\bm{\mathrm{w}}$ ranges have on the performance of agents?

During training, generalist agents are exposed to task weights within a predefined range in their observation space. In order to investigate the interpolation ability of our method, RQ3 studies the effect of altering the range of $\bm{\mathrm{w}}$ during training in Inspection3D. Until now, we have shown results for weights in $[0,1]$ (seen in SAC-S \textit{Gen}$[0,1]$ and SUSFAS \textit{Gen}$[0,1]$), which represents the full range of tasks containing $\bm{\mathrm{w}}$ combinations from $0$ to $1$. To these prior experiments, we add two generalists with incrementally narrowing ranges of $\bm{\mathrm{w}}$ during training, $[0.2,0.8]$ (reflected in SAC-S \textit{Gen}$[0.2,0.8]$ and SUSFAS \textit{Gen}$[0.2,0.8]$) and $[0.4,0.6]$ (reflected in SAC-S \textit{Gen}$[0.4,0.6]$ and SUSFAS \textit{Gen}$[0.4,0.6]$) in an effort the direct training toward tasks where $\bm{\mathrm{w}}$s are more equally distributed across $\phi$. If we consider cases where $\bm{\mathrm{w}}$ are equal, and given that there are three tunable weights in $\bm{\mathrm{w}}$, each weight setting will normalize to $0.33$. 

\begin{figure}[t]
  \begin{minipage}[c]{.49\linewidth}
     \centering
     \includegraphics[scale=0.98]{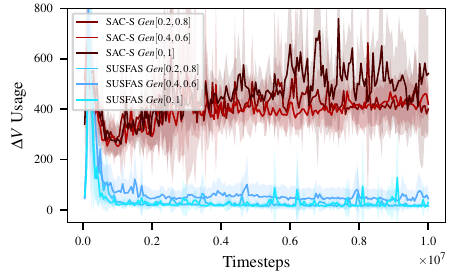} 
     \caption{RQ3 -- Inspection3D $\Delta V$ usage with varying generalist training ranges.}
     \label{fig:RQ3.2}
  \end{minipage}\hfill
  \begin{minipage}[c]{.49\linewidth}
    \centering
    \footnotesize	
    \begin{tabular}{lr}
    \toprule
    Label & \begin{tabular}{@{}c@{}}Normalized \\ AUC\end{tabular} \\
    \midrule
    SAC-S $Gen[0.4,0.6]$ & 0.83 \\
    SAC-S $Gen[0.2,0.8]$ & 0.90 \\
    SAC-S $Gen[0,1]$ & 1.00 \\
    \midrule
    SUSFAS $Gen[0.4,0.6]$ & 0.08 \\
    SUSFAS $Gen[0.2,0.8]$ & 0.00 \\
    SUSFAS $Gen[0,1]$ & \textbf{0.00} \\
    \bottomrule
    \end{tabular}
    \captionof{table}{RQ3 -- Ablation between specialists and generalists on Inspection3D}
    \label{tab:RQ3.2b}
  \end{minipage}
\end{figure}

From~\Cref{tab:RQ3.2b} displaying normalized AUC (lower is better) for $\Delta V$, we can see with more detail that narrowing the observed $\bm{\mathrm{w}}$ in SAC-S \textit{Gen} agents increases performance;  SAC-S $Gen[0.4,0.6]$ has a lower AUC than SAC-S $Gen[0,1]$. This describes the inability of SAC-S generalists to generalize across the range of tasks that Inspection3D presents. The SUSFAS generalists exhibit the opposite trend. Narrowing $\bm{\mathrm{w}}$ range in training to [0.4,0.6] increases normalized AUC for $\Delta V$ usage (e.g., SUSFAS Gen[0.4, 0.6] AUC = 0.08). This shows a promising trend in our method where a wide range of tasks (e.g., SUSFAS Gen[0.2, 0.8] with AUC = 0, and SUSFAS Gen[0, 1] with AUC = 0) produce the most capable agents as measured by normalized AUC for $\Delta V$ usage.

\paragraph{RA3:} Shown in~\Cref{tab:RQ3.2b} and~\Cref{fig:RQ3.2}, decreasing the ranges of the trained reward weights increases fuel usage (as proxied by $\Delta V$) for SUSFAS agents and decreases fuel usage for SAC-S agents on the Inspection3D environment. SUSFAS generalist agents trained with a wide band of reward weights perform better by maintaining a low $\Delta V$ throughout training. 

\section{Conclusion}

\textit{Successor features} (SFs) provide the inspiration behind a wide array of multi-task research. Our research refocuses SF prediction on problem spaces where safety plays a critical role. To that end, this work shows the importance of our expert stacking architecture and how it interacts favorably with an RTA controller. We show that RTA is required to see meaningful improvement in our SUSFAS method over SAC-S baselines.
We also study the effect of training progressively specialized agents to demonstrate the advantages of generalized training.

\subsection{Future Work and Limitations}

We did not explore a variety of structural changes, such as using a single shared encoder for the actions, weights, and observations across all SF predictors, nor did we explore the effect of using split encoders for the actor. A natural complement to successor features is \textit{predecessor features}~\citep{etlambda}, and it would be interesting to combine the two representations.

It would be interesting to investigate disentangling a goal from its representation in weight space, as was done in \citet{usf} and similar to \citet{mozifian_USFAforCC_2021}. Additionally, the nature of the underlying task space is not well understood. Is it possible that the potential collinearity of the dimensions of $\wt$ affects learning? Perhaps some form of regularization or enforcing of orthonormality in the weight feature space could positively impact the behavior and generalization of the agent.

Future work could also explore the nature of the environment and RTA controller in terms of the ability to incorporate the safety constraints from multiple controllers into the behavior. It would be interesting to better understand how the policy behavior might differ when the action inputs are more sparse, such as in a space environment that seeks to maximize the use of natural motion trajectories.

\subsubsection*{Broader Impact Statement}
\label{sec:broaderImpact}
While our work recognizes the need for safe operation of control systems when using reinforcement learning for real-world problems, much more understanding is needed and care must be taken to ensure appropriate deployment of these methods as part of any safety-critical system. We hope that our work can help practitioners make more informed decisions on safe operations for RL deployment. 

\bibliography{main_arxiv}
\bibliographystyle{rlc}

\appendix
\section{Appendix}

\subsection{Environments Details}

In Inspection3D, each observation consists of position $\varphi$, velocity $\dot\varphi$, sun angle $\theta_S$ and inspected points data so that $s = \{\varphi, \dot \varphi, \theta_S, P_i, P_c\}$, where $P_i$ is the cumulative number of inspected points and $P_c$ is a unit vector pointing toward the largest cluster of remaining uninspected points. The actions available in the environment are simply forces applied in each of the three principal axes. Each episode terminates after either a fixed number of maximum timesteps or when a defined minimum threshold $\tau \in [0, 100]$ of points inspected is reached. Our reward function is slightly modified from \citet{van2023deep} and consists of five components linearly weighted: $R = w_OR_O + w_CR_C + w_TR_T + w_vR_{\Delta v} + w_AR_{A}$ where each is component consists of the following.

\begin{enumerate}
    \item Observed points $R_{O}(s_t, a_t, s_{t+1}) = 0.01\left([P_i(t+1) - P_i(t)] + \chi_{[\tau,100]}(P_i)\right)$, based on the number of new points observed by the agent at timestep $t$ along with a termination reward if the minimum number of points required have been inspected.
    \item Crash reward (penalty) $R_{C} = -1$, if agent exits inspection area or crashes into target, otherwise $R_{C} = 0$.
    \item Timestep (penalty) $R_T(s_t, a_t, s_{t+1}) = -0.0001$, a constant penalty encouraging the agent to solve the task quickly.
    \item Fuel usage (penalty) $R_{\Delta \vec{v}} = - \sum_{t} \Delta \vec{v}_t$, a cumulative penalty for the total amount of fuel used during the episode.
    \item RTA (penalty) $R_{A} = - 0.01n_{A}$, where $n_{A}$ is number of timesteps where the RTA was active.
\end{enumerate}

The agent models a ``deputy'' spacecraft, which is orbiting and attempting to inspect a passive ``chief'' satellite modeled as a spherical point cloud containing $100$ points. Translational motion is governed by the Clohessy--Wiltshire linearized dynamics in Hill's frame (i.e., centered on the chief satellite), and the rotational position (attitude) of the spacecraft is adjusted by the simulator so the orientation of the observation sensors are always directed towards the chief. Points are considered inspected when they fall within a defined perception cone and are currently illuminated by the Sun. The environment includes a suite of RTA filters that intercept and minimally modify potentially unsafe actions (e.g., actions that would cause a crash or violate dynamic speed constraints) based on {\em Active Set Invariance Filtering} (ASIF)~\citep{ASIF_2018}.

\subsection{Hyperparameters}

\Cref{fig:app:hparams} lists hyperparameters used in the SAC, CUSFAS, and SUSFA algorithms. Notably, CUSFAS and SUSFA algorithms use hyperparameters most commonly used in the SAC algorithm, the only difference between the two being the number of stacked Successor Feature Approximator (SFA) networks used for the critic.

\begin{table}[h]
\centering
\begin{tabular}{lllll}
\toprule
Parameter                                 & SAC               & CUSFAS            & SUSFA             \\
\midrule
Optimizer                                 & Adam              & Adam              & Adam              \\
Learning rate                             & $3\text{e-}4$     & $3\text{e-}4$     & $3\text{e-}4$     \\
Discount ($\gamma$)                       & 0.99              & 0.99              & 0.99              \\
Replay buffer size                        & $1\text{e}6$      & $1\text{e}6$      & $1\text{e}6$      \\
Number of hidden layers (all networks)    & 2                 & 2                 & 2                 \\
Number of hidden units per layer (actor)  & 256               & 256               & 256               \\
Number of hidden units per layer (critic) & 256               & 256               & 256               \\
Number of samples per minibatch           & 256               & 256               & 256               \\
Entropy target                            & $-\text{dim}(\A)$ & $-\text{dim}(\A)$ & $-\text{dim}(\A)$ \\
Nonlinearity                              & ReLU              & ReLU              & ReLU              \\
Target smoothing coefficient ($\eta$)     & 0.005             & 0.005             & 0.005             \\
Target update interval                    & 1                 & 1                 & 1                 \\
Gradient steps ($k$)                           & 1                 & 1                 & 1                 \\
Number of stacked SFA Networks            & $-$               & 1                 & $d$                 
\end{tabular}
\captionof{table}{SAC, CUSFAS, and SUSFAS Hyperparameters, where $d$ is the number of SFs.}
\label{fig:app:hparams}
\end{table}

\end{document}